\newcommand*{\CVPR}{}

\newcommand*{\CAMREADY}{}

\ifdefined\NIPS
	\documentclass{article}
	\usepackage{nips15submit_e,times}
	\usepackage{hyperref}
	\usepackage{url}
\fi
\ifdefined\CVPR
	\documentclass[10pt,twocolumn,letterpaper]{article}
	\usepackage{cvpr}
	\usepackage{times}
	\usepackage{epsfig}
	\usepackage{graphicx}
	\usepackage{amsmath}
	\usepackage{amssymb}
\fi
\ifdefined\AISTATS
	\documentclass[twoside]{article}
	\ifdefined\CAMREADY
		\usepackage[accepted]{aistats2015}
	\else
		\usepackage{aistats2015}
	\fi
	\usepackage{url}
\fi

\usepackage{color}
\usepackage{amsfonts}
\usepackage{amsmath}
\usepackage{algorithm}
\usepackage{algorithmic}
\usepackage{graphicx}
\usepackage{bbm}
\usepackage{amsthm}
\usepackage{wrapfig}
\ifdefined\NIPS
	\usepackage[square,comma,numbers]{natbib}
\fi
\ifdefined\CVPR
	\usepackage[square,comma,numbers]{natbib}
\fi
\ifdefined\AISTATS
	\usepackage[round,authoryear]{natbib}
\fi

\newtheorem{theorem}{Theorem}

\def\be{\begin{equation}}
\def\ee{\end{equation}}
\def\beas{\begin{eqnarray*}}
\def\eeas{\end{eqnarray*}}
\def\bea{\begin{eqnarray}}
\def\eea{\end{eqnarray}}

\newcommand{\x}{{\mathbf x}}
\newcommand{\y}{{\mathbf y}}
\newcommand{\z}{{\mathbf z}}
\newcommand{\uu}{{\mathbf u}}

\newcommand{\w}{{\mathbf w}}

\newcommand{\bb}{{\mathbf b}}
\newcommand{\cc}{{\mathbf c}}
\newcommand{\dd}{{\mathbf d}}

\newcommand{\R}{{\mathbb R}}
\newcommand{\N}{{\mathbb N}}
\newcommand{\1}{{\mathbf 1}}

\newcommand{\abs}[1]{\lvert#1 \rvert}

\newcommand{\inprod}[2]  {\left\langle{#1},{#2}\right\rangle}
\DeclareMathOperator*{\argmax}{argmax}

\newcommand{\mexu}[2]  {\underset{#2}{MEX_{#1}}}

\ifdefined\CAMREADY
	\newcommand{\simnetsref}{cohen2014simnets}
\else
	\newcommand{\simnetsref}{anonymous}
\fi

\ifdefined\CVPR
	\usepackage[breaklinks=true,bookmarks=false]{hyperref}
	\ifdefined\CAMREADY
		\cvprfinalcopy 
	\fi
	\def\cvprPaperID{1464} 
	
\fi
\ifdefined\NIPS

	\ifdefined\CAMREADY
		\nipsfinalcopy
	\fi
\fi

\begin{document}

\ifdefined\NIPS
	\title{Deep SimNets}
	\author{
	Nadav Cohen \\
	The Hebrew University of Jerusalem \\
	\texttt{cohennadav@cs.huji.ac.il} \\
	\And 
	Or Sharir \\
	The Hebrew University of Jerusalem \\
	\texttt{or.sharir@cs.huji.ac.il} \\
	\And 
	Amnon Shashua \\
	The Hebrew University of Jerusalem \\
	\texttt{shashua@cs.huji.ac.il} \\
	}
	\maketitle
\fi
\ifdefined\CVPR
	\title{Deep SimNets}
	\author{
	Nadav Cohen \\
	The Hebrew University of Jerusalem \\
	\texttt{cohennadav@cs.huji.ac.il} \\
	\and
	Or Sharir \\
	The Hebrew University of Jerusalem \\
	\texttt{or.sharir@cs.huji.ac.il} \\
	\and    
	Amnon Shashua \\
	The Hebrew University of Jerusalem \\
	\texttt{shashua@cs.huji.ac.il} \\
	}
	\maketitle
	\ifdefined\CAMREADY
	\fi
\fi
\ifdefined\AISTATS
	\twocolumn[
	\aistatstitle{Deep SimNets}
	\ifdefined\CAMREADY
		\aistatsauthor{Nadav Cohen \And Or Sharir \And Amnon Shashua}
		\aistatsaddress{The Hebrew University of Jerusalem \And The Hebrew University of Jerusalem \And The Hebrew University of Jerusalem}
	\else
		\aistatsauthor{Anonymous Author 1 \And Anonymous Author 2 \And Anonymous Author 3}
		\aistatsaddress{Unknown Institution 1 \And Unknown Institution 2 \And Unknown Institution 3}
	\fi
	]	
\fi

\begin{abstract}
We present a deep layered architecture that generalizes convolutional neural networks (ConvNets).  The architecture, called SimNets, is driven by two operators: (i) a similarity function that generalizes inner-product, and (ii) a log-mean-exp function called MEX that generalizes maximum and average.  The two operators applied in succession give rise to a standard neuron but in "feature space".  The feature spaces realized by SimNets depend on the choice of the similarity operator.  The simplest setting, which corresponds to a convolution, realizes the feature space of the Exponential kernel, while other settings realize feature spaces of more powerful kernels (Generalized Gaussian, which includes as special cases RBF and Laplacian), or even dynamically learned feature spaces (Generalized Multiple Kernel Learning).  As a result, the SimNet contains a higher abstraction level compared to a traditional ConvNet.  We argue that enhanced expressiveness is important when the networks are small due to run-time constraints (such as those imposed by mobile applications).  Empirical evaluation validates the superior expressiveness of SimNets, showing a significant gain in accuracy over ConvNets when computational resources at run-time are limited.  We also show that in large-scale settings, where computational complexity is less of a concern, the additional capacity of SimNets can be controlled with proper regularization, yielding accuracies comparable to state of the art ConvNets.
\end{abstract}

\section{Introduction} \label{sec:intro}
Deep neural networks, and convolutional neural networks (\emph{ConvNets}) in particular, have had a dramatic impact in advancing the state of the art in computer vision, speech analysis, and many other domains (cf.~\cite{krizhevsky2012imagenet,szegedy2014going,hannun2014deepspeech}).  It has been demonstrated time and time again, that when ConvNets are trained in an end-to-end manner, they deliver significantly better results than systems relying on manually engineered features.

The goal of this paper is to introduce a generalization of ConvNets we call Similarity Networks (\emph{SimNets}), that preserves the simplicity and effectiveness of ConvNets, yet has a \emph{higher abstraction level}.  In a nutshell, the inner-product operator, which lies at the core of the ConvNet architecture, is replaced by an inner-product in ``feature space''.  The feature spaces are controlled by a family of kernel functions which include in particular the conventional (linear) inner-product as a special case. 

We argue that the incentive for designing deep networks with a higher abstraction level than ConvNets, arises from the need for small networks that could fit into mobile platforms in terms of space and run-time.  With small networks the approximation error becomes a limiting factor, which could be ameliorated through network architectures that are based on a higher level of abstraction.

The SimNet architecture is based on two operators.  The first is analogous to, and generalizes, the inner-product operator of neural networks.  The second, as special cases, plays the role of non-linear activation and pooling, but has additional capabilities that take SimNets far beyond ConvNets.  In a detailed set of experiments, the SimNet architecture achieves state of the art accuracy using networks with complexity comparable to that of top performing ConvNets.  However, when network complexity is limited, SimNets deliver a significant boost in accuracy.

Recently, the task of reducing run-time complexity of ConvNets is receiving increased attention.  For example, a method named FitNets~(\cite{Romero:2014tg}), based on the knowledge distillation principle~(\cite{hinton2014distilling}), has been suggested in order to assist in compressing deep networks.  In~\cite{Srivastava:2015uq}, a form of gating inspired by Long Short-Term Memory recurrent networks is introduced, allowing training of very deep and narrow networks.  Another line of work considers imposing structural constrains on network weights, such as sparsity, in order to improve run-time efficiency (\cite{Figurnov:2015wj,Collins:2014vy,Han:2015vn,Chen:2015wg,Chen:2015wz}).  Alternatively, network weights may be factorized using matrix or tensor decompositions, reducing storage and computational complexity, at the expense of marginal deterioration in accuracy (\cite{Denton:2014vs,Jaderberg:2014ul,Zhang:2015ey,Lebedev:2014vb,Novikov:2015uq,Yang:2015vd,Cheng:2015vd}).  All of these approaches consider ConvNets (or neural networks) as a baseline, and use supplementary techniques to reduce run-time complexity.  In this work, we propose the alternative (generalized) SimNet architecture, and argue that it is inherently more efficient than ConvNets.  The techniques listed here for reducing run-time complexity of ConvNets could just as well be applied to SimNets, thereby resulting in even more computationally efficient models.

\section{The SimNet architecture} \label{sec:simnet_arch}
A feed-forward fully-connected neural network, also known as a multilayer perceptron (MLP), is based on a single operator.  Given $\x\in\R^d$ as input to a layer of neurons, the output of the $r$'th neuron in the layer is $\sigma(\w_r^\top\x+b_r)$, where $\sigma(\cdot)$ is a non-linear activation function.  An MLP is constructed by forward chaining the input/output operation to create a layered network.  The learned parameters of the network are the weight vectors $\w_r$ and biases $b_r$, per neuron.

The SimNet architecture consists of two operators.  The first operator is a weighted similarity function between an input $\x\in\R^d$ and a template $\z\in\R^d$:
$${\rm similarity~operator:~} \uu^\top\phi(\x,\z)$$
where $\uu\in\R_+^d$ is a weight vector and $\phi:\R^d\times\R^d\to\R^d$ is a point-wise similarity mapping.  We consider two forms of similarity mappings: the ``linear'' form $\phi_{lin}(\x,\z)_i=x_i z_i$, and the ``$\ell_p$'' form $\phi_{\ell_p}(\x,\z)_i=-\abs{x_i-z_i}^p$ defined for $p>0$.  Note that when setting $\uu=\1$, the corresponding similarities reduce to inner-product and $p$-distance (by the power of $p$) respectively.  Note also that unlike the MLP operator, the similarity does not include a bias term.  This functionality is covered, in a much more general sense, by the second operator described below.

For the second SimNet operator we define \emph{MEX} -- a log-mean-exp function:
\be
\mexu{\beta}{i=1,...,n}\{c_i\}:=\frac{1}{\beta}\log\left(\frac{1}{n}\sum_{i=1}^n\exp\{\beta{\cdot}c_i\}\right)
\label{eqn:mex_def}
\ee
The parameter $\beta\in\R$ spans a continuum between maximum ($\beta\to+\infty$), average ($\beta\to0$) and minimum ($\beta\to-\infty$), and for a fixed value of $\beta$ the function is smooth and exhibits the following ``collapsing'' property \footnote{The collapsing property, as well as smoothly generalizing maximum and average, will prove to be essential for us.  We are not aware of other functions that meet these three requirements.  Specifically, the common softmax function $\frac{1}{\beta}\log\left(\sum_{i}\exp\{\beta{\cdot}c_i\}\right)$ collapses and generalizes maximum but does not generalize average, and the alternative softmax function ${\sum_{i}c_{i}e^{\beta c_i}}/{\sum_{i}e^{\beta c_i}}$ generalizes maximum and average but does not collapse.}:
\beas
&&MEX_\beta\{ MEX_\beta\{c_{ij}\}_{1\leq j\leq m}\}_{1\leq i\leq n} \\
&&~~~~~=MEX_\beta\{c_{ij}\}_{1\leq j\leq m,1\leq i\leq n}
\eeas

Given the definition in eqn.~\ref{eqn:mex_def}, the second SimNet operator consists of taking MEX over an input $\x\in\R^d$ with a bias vector $\bb\in\R^d$ -- one per input coordinate \footnote{The MEX operator can be viewed as an ``inner-product in log-space''.  More accurately, if $\x$ and $\bb$ are log-space representations of two vectors $\cc$ and $\dd$ respectively (i.e. $x_i=\log{c_i}$ and $b_i=\log{d_i}$), then $MEX_{\beta=1}\{x_i + b_i\}_{i}=\log\inprod{\cc}{\dd}-\log{d}$.  In words, the MEX operator (with $\beta=1$) taken over the log-space representations of $\cc$ and $\dd$ is equal (up to an additive constant) to the log-space representation of their inner-product.}: 
$${\rm MEX~operator:~} MEX_{\beta>0}\{x_i + b_i\}_{i=1,..,d}$$
Note that unlike a conventional MLP unit which has a bias scalar, a MEX unit has a vector of biases.  We may choose to omit part or all of the biases as part of a network design.  For example, when all biases are dropped the MEX operator implements a soft trade-off between maximum and average.

\section{SimNet MLP} \label{sec:simnet_mlp}
\begin{figure*}
\includegraphics[width=\textwidth,height=\textheight,keepaspectratio]{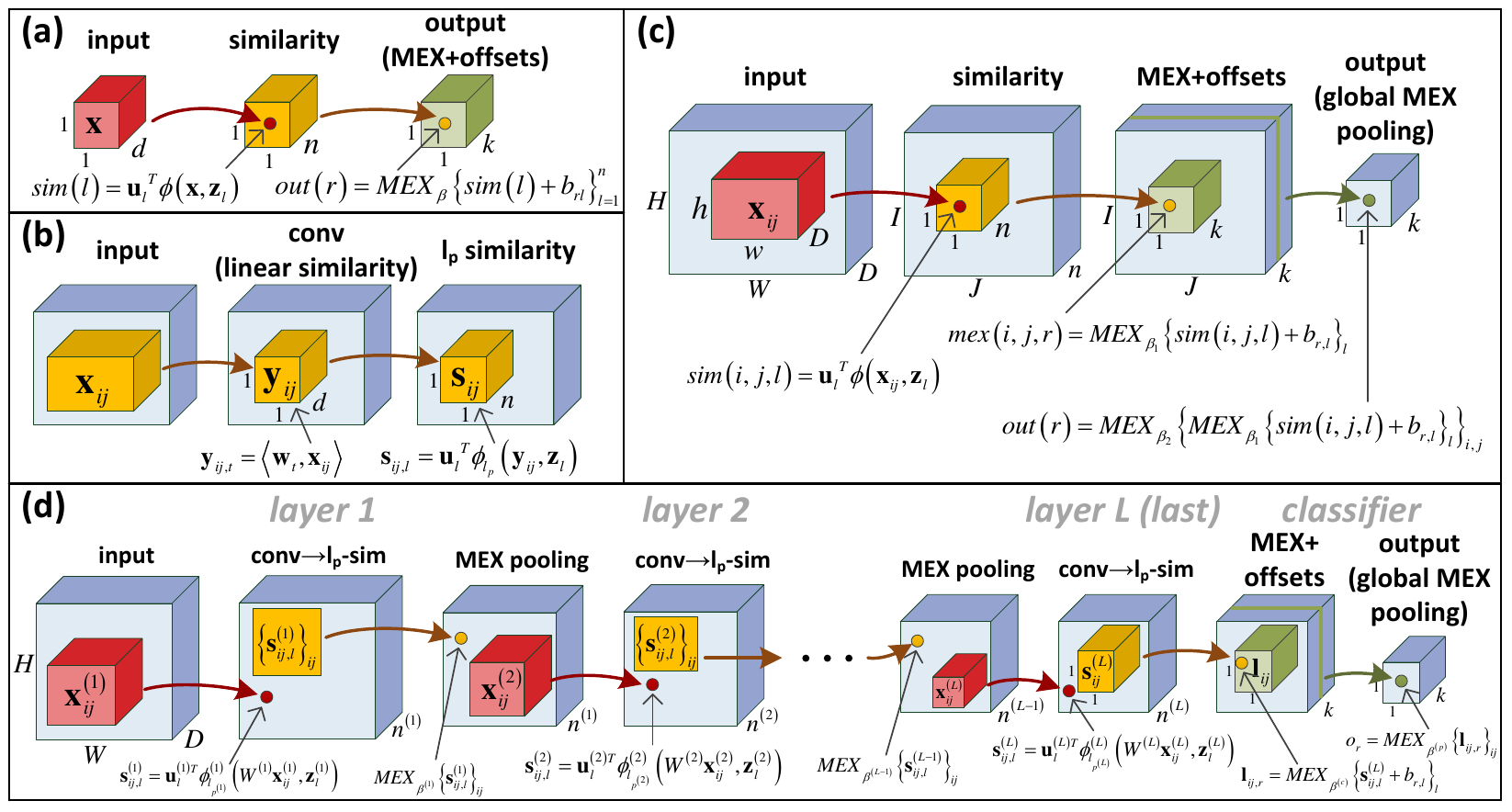}
\caption{\footnotesize{(a)~SimNet MLP~--~SimNet analogy of MLP with single hidden layer (sec.~\ref{sec:simnet_mlp})~~(b)~conv$\to\ell_p$-sim structure~--~implements whitened $\ell_p$ similarity (sec.~\ref{subsec:whiten})~~(c)~SimNet MLPConv~--~single layer SimNet for processing images (sec.~\ref{subsec:simnet_mlpconv})~~(d)~L-layer SimNet for processing images (sec.~\ref{subsec:going_deep}).  Best viewed in color.}}
\label{fig:arch_schemes}
\vspace{-3mm}
\end{figure*}

A SimNet analogy of an MLP with a single hidden layer is obtained by applying the two operators defined in sec.~\ref{sec:simnet_arch} one after the other -- similarity followed by MEX.  The resulting network is illustrated in fig.~\ref{fig:arch_schemes}(a).  It includes $n$ hidden similarity units corresponding to weighted templates $\{(\z_l,\uu_l)\}_{l=1}^n$, and $k$ output MEX units associated with bias vectors $\{\bb_r\}_{r=1}^k$.  Denote by $h_r(\x)$ the value of the $r$'th output unit when the network is fed with input $\x\in\R^d$: $h_r(\x):=MEX_{\beta}\{\uu_l^\top\phi(\x,\z_l)+b_{rl}\}_{l=1}^n$ \footnote{Note that with uniform weights ($\uu_l\equiv\1$), linear similarity mapping $\phi$ and $\beta\to+\infty$ we have $h_r(\x)=\max\left\{\z_l^\top\x+b_{rl}\right\}_{l=1}^n$, i.e. the network outputs are ``maxout'' units~(\cite{goodfellow2013maxout}).  SimNet MLP is not the first to generalize maxout.  Other generalizations have been suggested, notably the recently proposed $L_p$ unit~(\cite{gulcehre2014learned}), which is defined by $(\sum_l |\z_l^\top\x+b_{rl}|^p)^{1/p}$, and tends to $\max_l\left\{\abs{\z_l^\top\x+b_{rl}}\right\}$ as $p\to+\infty$.  The differences between SimNet MLP and $L_p$ unit as maxout generalizations are: (i) $L_p$ unit generalizes maximum of absolute values which only coincides with maxout if the arguments are non-negative, and (ii) $L_p$ unit tries to realize maxout with a single operator whereas SimNet MLP implements maxout with a succession of two operators.}.  As a classifier of $\x$ into one of $k$ categories, the network predicts the label $r$ for which $h_r(\x)$ is maximal:
$$\hat{y}(\x)=\argmax_{r=1,...,k}MEX_\beta\{\uu_l^\top\phi(\x,\z_l)+b_{rl}\}_{l=1}^n$$

As it turns out, SimNet MLP is closely related to kernel machines.  In particular, with linear similarity, i.e. with the inner-product operator on which neural networks are based, it is a support vector machine (SVM) based on the Exponential kernel.  Replacing the linear similarity with $\ell_p$ boosts the abstraction level of SimNet MLP, by lifting it to a Generalized Multiple Kernel Learning (GMKL,~\cite{varma2009more}) engine with a Generalized Gaussian kernel.  The remainder of this section provides the details.

SimNet MLP outputs can be written as: 
\beas
h_r(\x)&=&MEX_{\beta}\{\uu_l^\top\phi(\x,\z_l)+b_{rl}\}_{l=1}^n \\
&=&\frac{1}{\beta}\ln \left(\frac{1}{n}\sum_{l=1}^{n}\alpha_{rl}\exp\left\{\beta\sum_{i=1}^{d}u_{l,i}\phi(\x,\z_l)_i\right\}\right) \\
&=&\sigma\left(\sum_{l=1}^n\alpha_{rl}\cdot K_\theta (\x,\z_l)\right)
\eeas
where $\alpha_{rl}:=\exp\{\beta b_{rl}\}$, $\theta=(\phi,\uu)$, and $\sigma(t)=(1/\beta)\ln(t/n)$ is a non-linear activation function.  The mapping $K_\theta$ for the linear and $\ell_p$ similarities takes the following forms:
\beas
K_{lin}(\x,\z)&=&\exp\left\{\beta\x^\top\z \right\} \\
K_{\ell_p}(\x,\z_l)&=&\exp\left\{-\beta \sum_{i=1}^{d}u_{l,i}\abs{x_i-z_{l,i}}^p\right\}
\eeas
$K_{lin}$ is known as the \emph{Exponential}  kernel~(\cite{scholkopf2002learning}), and $K_{\ell_p}$ is a GMKL.  Specifically, fixing uniform weights ($\uu_l\equiv\1$) and $p\leq2$ reduces $K_{\ell_p}$ to what is known as the \emph{Generalized Gaussian} kernel.  For the particular cases $p=2$ and $p=1$ we get the radial basis function (RBF) and Laplacian kernels respectively.  When the weights $\uu_l$ and/or order $p$ are learned, the exact underlying kernel is selected during training and we amount at a GMKL.

Denoting by $\psi_\theta$ a feature mapping associated with $K_\theta$, we get: 
$$h_r(\x)=\sigma\left(\inprod{\psi_\theta(\x)}{\w_r}\right)$$
where $\w_r:=\sum_{l=1}^n\alpha_{rl}\psi_\theta(\z_l)$ is a learned vector in feature space.  We thus conclude that SimNet MLP output units are ``neurons in feature space'', where the space corresponds to the Exponential kernel in the case of linear similarity, and to the Generalized Gaussian kernel in the case of $\ell_p$ similarity with fixed weights $\uu_l$ and order $p$.  When the weights and/or order are learned, the feature space is selected during training, which is equivalent to saying that SimNet MLP is a GMKL.

One may ask if perhaps a different choice of kernel, more elaborate than Generalized Gaussian, will suffice in order to capture SimNet MLP with $\ell_p$ similarity and learned weights as a simple kernel machine.  Apparently, as theorem~\ref{thm} (proven in~\cite{\simnetsref}) shows, such a kernel does not exist, i.e. a GMKL is indeed necessary in order to represent SimNet MLP in all its glory. 
\begin{theorem}
\label{thm}
For any dimension $d\in\N$, and constants $c>0$ and $p>0$, there are no mappings $Z:\R^d\to\R^d$ and $U:\R^d\to\R_+^d$ and a kernel $K:(\R^d\times\R_+^d)\times(\R^d\times\R_+^d)\to\R^d\times\R_+^d$, such that for all $\z,\x\in\R^d$ and $\uu\in\R_+^d$: $K\left([Z(\x),U(\x)],[\z,\uu]\right)=\exp\left\{-c\sum_{i=1}^d u_i\abs{x_i-z_i}^p\right\}$.
\end{theorem}

\section{Deep SimNets for processing images} \label{sec:deep_simnets}
In the previous section we presented the basic MLP version of SimNets.  In this section we describe two (orthogonal) directions of extension.  The first is the addition of locality, sharing and pooling for processing images (SimNet MLPConv, sec.~\ref{subsec:simnet_mlpconv}), while the second focuses on deepening the network (adding layers) for enhanced capacity (sec.~\ref{subsec:going_deep}).  In this context we introduce a ``whitened'' $\ell_p$ similarity layer through a succession of a convolution (linear similarity) followed by $\ell_p$ similarity with receptive field $1\times1$.  
\subsection{SimNet MLPConv} \label{subsec:simnet_mlpconv}
The extension of SimNet MLP for processing images follows the line of the MLPConv structure suggested in~\cite{lin2013network}, and we accordingly refer to it as SimNet MLPConv.  In particular, \cite{lin2013network} convolved a standard MLP across an incoming 3D array by successively applying it to patches and stacking the outputs in a spatially coherent manner.  This results in a bank of feature maps, which may be summarized into prediction scores through global average pooling.  SimNet MLPConv follows the same principles -- a SimNet MLP is convolved across an incoming 3D array, and the resulting feature maps are summarized via global MEX pooling.  An illustration of SimNet MLPConv is provided in fig.~\ref{fig:arch_schemes}(c).  In the figure, $\x_{ij}\in\R^{hwD}$ refers to the input patch in location ${ij}$, $\z_l\in\R^{hwD}$ and $\uu_l\in\R^{hwD}_+$ denote similarity templates and weights respectively, $\phi:\R^{hwD}\times\R^{hwD}\to\R^{hwD}$ is the similarity mapping (linear or $\ell_p$), $\beta_1\in\R$ and $b_{rl}\in\R$ are the MEX parameter and offsets of the underlying SimNet MLP, and $\beta_2\in\R$ is the MEX parameter of the final global pooling layer.

When used to classify images, the prediction rule associated with SimNet MLPConv is given by: $\hat{y}(input)=\argmax_{r}MEX_{\beta_2}\left\{MEX_{\beta_1}\left\{\uu_l^\top\phi(\x_{ij},\z_l)+b_{r,l}\right\}_{l}\right\}_{i,j}$.  Setting $\beta_1=\beta_2=\beta$, and using the collapsing property of MEX, we get a ``patch-based'' version of SimNet MLP's classification: 
$$\hat{y}(input)=\argmax_{r}\mexu{\beta}{i,j,l}\{\uu_l^\top\phi(\x_{ij},\z_l)+b_{r,l}\}$$
It can be shown~(\cite{\simnetsref}) that all results put forth in sec.~\ref{sec:simnet_mlp} for relating SimNet MLP to kernel machines apply to SimNet MLPConv as well, but with the underlying kernels being based on ``patch-representations''.  In other words, SimNet MLPConv -- a ``patch-based'' extension of SimNet MLP, maintains all kernel relations of the latter, with a ``patch-based'' extension of the  underlying kernels.
\subsection{Whitening with convolutional layer} \label{subsec:whiten}
We now describe a simple yet powerful addition to the $\ell_p$ similarity operator.  Recall that the $\ell_p$ similarity between an input $\x\in\R^d$ and a template $\z\in\R^d$ with weights $\uu\in\R^d_+$, is defined by $-\sum_{i=1}^d u_i\abs{x_i-z_i}^p$.  Up to a constant that depends on $\uu$ (and $p$), this is equal to the log probability density of the input $\x$ being drawn from a Generalized Gaussian distribution with \emph{independent components}, shape $p$, mean $\z$, and scales $\uu^{-1/p}$.  These ideas are further developed in sec.~\ref{sec:pre_train}, however it is clear at this point that in order to capture this probabilistic model, it would be desirable for the input $\x$ to have statistically independent coordinates.  Common practice in such cases is to seek for a matrix $W$ for which the linearly transformed vector $W\x$ has independent coordinates.  This is referred to in the literature as ICA -- independent component analysis~(\cite{hyvarinen2004independent}).  Assuming such a matrix is found, it would then be natural to ``whiten'' inputs, i.e. multiply them by $W$, before measuring their $\ell_p$ similarities to weighted templates.  Besides better compliance with the coordinate independence assumption, this also gives rise to the possibility of dimensionality reduction.  In particular, we may set the matrix $W$ to cancel-out low-variance principal components of $\x$, thereby producing whitened vectors of a lower dimension.  This can be useful for both noise reduction and computational efficiency.

In the context of SimNet MLPConv, adding support for whitening before $\ell_p$ similarity is simple -- it merely requires a convolutional layer (linear similarity) followed by an $\ell_p$ similarity layer with receptive field $1\times1$.  Such a construct, which we refer to as conv$\to\ell_p$-sim, is illustrated in fig.~\ref{fig:arch_schemes}(b).  In this figure, input patches $\x_{ij}$ are transformed into $d$-dimensional vectors $\y_{ij}$ by a convolutional layer with $d$ filters $\w_t$ that hold the rows of the whitening matrix $W$.  The whitened vectors $\y_{ij}$ are then matched against $n$ weighted templates in the $\ell_p$ similarity layer, producing $n$ similarity maps as output.  To recap, one may add whitening to $\ell_p$ similarity by replacing the similarity layer with a conv$\to\ell_p$-sim structure, which consists of convolution followed by $1\times1$ similarity. 

In sec.~\ref{sec:pre_train} we describe how to pre-train a conv$\to\ell_p$-sim structure, and in particular how to initialize the filters so that they perform the whitening transformation they are intended for.  Before that however, we show how SimNet MLPConv can be extended into an image processing SimNet of arbitrary depth.
\subsection{Going deep with SimNet MLPConv} \label{subsec:going_deep}
After laying out the basic SimNet construct (SimNet MLP~--~sec.~\ref{sec:simnet_mlp}), equipping it with spatial structure (SimNet MLPConv~--~sec.~\ref{subsec:simnet_mlpconv}), and adding whitening to its $\ell_p$ similarity (conv$\to\ell_p$-sim~--~sec.~\ref{subsec:whiten}), we are finally in a position to define an arbitrarily deep SimNet for processing images.  Our starting point is SimNet MLPConv with whitened $\ell_p$ similarity.  This network accounts for a single layer (conv$\to\ell_p$-sim) followed by a classifier (classification MEX and global MEX pooling).  Adding depth to the network simply amounts to appending preceding conv$\to\ell_p$-sim layers, optionally separated by MEX pooling.  A general L-layer SimNet following this architectural prescription is illustrated in fig.~\ref{fig:arch_schemes}(d).  In this structure, conv$\to\ell_p$-sim layers measure whitened $\ell_p$ similarities of incoming patches to weighted templates, MEX pooling operations summarize spatial regions in similarity maps by MEX'ing them together (note that both average pooling and max pooling are special cases of this), the MEX classification uses its offsets $b_{rl}$ to classify each location in the final similarity maps, and the final global MEX pooling summarizes the local classifications into global class scores.  The parameters that may be learned during training are: 
$W^{(1)}{\ldots}W^{(L)}$~--~linear filters in conv$\to\ell_p$-sim;
$\z^{(1)}_l{\ldots}\z^{(L)}_l$ and $\uu^{(1)}_l{\ldots}\uu^{(L)}_l$~--~similarity templates and weights  in conv$\to\ell_p$-sim;
$p^{(1)}{\ldots}p^{(L)}$~--~similarity orders in conv$\to\ell_p$-sim;
$\beta^{(1)}{\ldots}\beta^{(L)}$~--~MEX parameters in local pooling;
$\beta^{(c)}$~--~MEX parameter in classification;
$b_{rl}$~--~MEX offsets in classification;
$\beta^{(p)}$~--~MEX parameter in global pooling.
In the following section we describe methods for initializing these parameters prior to training (pre-training).

\section{Pre-training} \label{sec:pre_train}
In this section we briefly describe a method for pre-training an L-layer SimNet as illustrated in fig.~\ref{fig:arch_schemes}(d).  Our initialization scheme covers the parameters of conv$\to\ell_p$-sim layers (linear filters $W^{(1)},...,W^{(L)}$, similarity templates $\z^{(1)}_l,...,\z^{(L)}_l$, weights $\uu^{(1)}_l,...,\uu^{(L)}_l$ and orders $p^{(1)},...,p^{(L)}$), assuming predetermined local MEX pooling parameters ($\beta^{(1)},...,\beta^{(L)}$).  Two attractive properties of the scheme are: (i) it is unsupervised (does not require any labels), and (ii) it gives rise to automatic selection of the number of channels in the convolutions and similarities of conv$\to\ell_p$-sim layers.

The initialization is applied layer by layer in a forward sweep, thus in order for it to be defined, it suffices to consider a single conv$\to\ell_p$-sim layer (fig.~\ref{fig:arch_schemes}(b)).  Recall from sec.~\ref{subsec:whiten} that we interpret the convolution in conv$\to\ell_p$-sim as a linear transformation that whitens (and possibly reduces the dimension of) input patches prior to similarity measurements.  Accordingly, we initialize its filters $\w_1,...,\w_d$ as the rows of a whitening matrix $W$ estimated via ICA~(\cite{hyvarinen2004independent}) on patches.  

Turning to the initialization of similarity templates ($\z_1,...\z_n$), weights ($\uu_1,...,\uu_n$) and order ($p$), we recall that an $\ell_p$ similarity between an input $\y\in\R^d$ and a template $\z\in\R^d$ with weights $\uu\in\R^d_+$, is defined to be $-\sum_{t=1}^d u_t\abs{y_t-z_t}^p$.  Consider now a probability distribution over $\R^d$ defined by a mixture of $n$ Generalized Gaussians (with priors $\lambda_l\geq0,\sum_l\lambda_l=1$), all having the same shape parameter ($\beta>0$), and each having independent coordinates with separate scales and means ($\alpha_{l,t}>0$ and $\mu_{l,t}\in\R$ respectively, for coordinate $t$ of component $l$): 
$$P(\y)=\sum_{l=1}^n\lambda_l\prod_{t=1}^d\frac{\beta}{2\alpha_{l,t}\Gamma(1/\beta)}e^{-\left(\abs{y_t-\mu_{l,t}}/\alpha_{l,t}\right)^{\beta}}$$
The log probability density of a vector drawn from this distribution being equal to $\y$ and originating from component $l$ is: $\log P(\y\land \text{comp.}~l)=-\sum_{t=1}^d\alpha_{l,t}^{-\beta}\abs{y_t-\mu_{l,t}}^{\beta}+c_l$, where $c_l:=\log\left\{\lambda_l\prod_{t=1}^d\frac{\beta}{2\alpha_{l,t}\Gamma(1/\beta)}\right\}$ is a constant that does not depend on $\y$.  This implies that if we model whitened patches $\y_{ij}$ with a Generalized Gaussian mixture as above, initializing the similarity templates via $z_{l,t}=\mu_{l,t}$, the weights via $u_{l,t}=\alpha_{l,t}^{-\beta}$ and the order via $p=\beta$ would give: 
$$\uu_l^\top\phi_{\ell_p}(\y_{ij},\z_l)=\log P(\y\land \text{comp.}~l)-c_l$$
In words, similarity channel $l$ would hold, up to a constant, the probabilistic heat map of component $l$ and the whitened patches $\y_{ij}$.  This observation suggests estimating the parameters of the mixture (shape $\beta$, scales $\alpha_{l,t}$ and means $\mu_{l,t}$) based on whitened patches (via EM, cf.~\cite{bazi2007image}), and initializing the similarity parameters accordingly.  We note in passing that it is possible to append additive biases $b_l$ to the similarity (through offsets of the succeeding MEX operator), in which case initializing these via $b_l=c_l$ would make the probabilistic heat maps exact (not up to a constant).

Finally, as stated above, the initialization scheme presented induces an automatic selection of the number of convolution and similarity channels in conv$\to\ell_p$-sim.  The number of convolution channels corresponds to the dimension to which input patches are reduced during whitening, thus may be set via methods for estimating effective dimensionality of data (e.g.~\cite{seghouane2007bayesian}).  Similarity channels correspond to components in the mixture estimated for whitened patches, thus may be set via methods for estimating the number of components in a mixture (e.g.~\cite{celeux1996entropy}).

\section{Experiments}
\begin{figure*}
\includegraphics[width=\textwidth,height=\textheight,keepaspectratio]{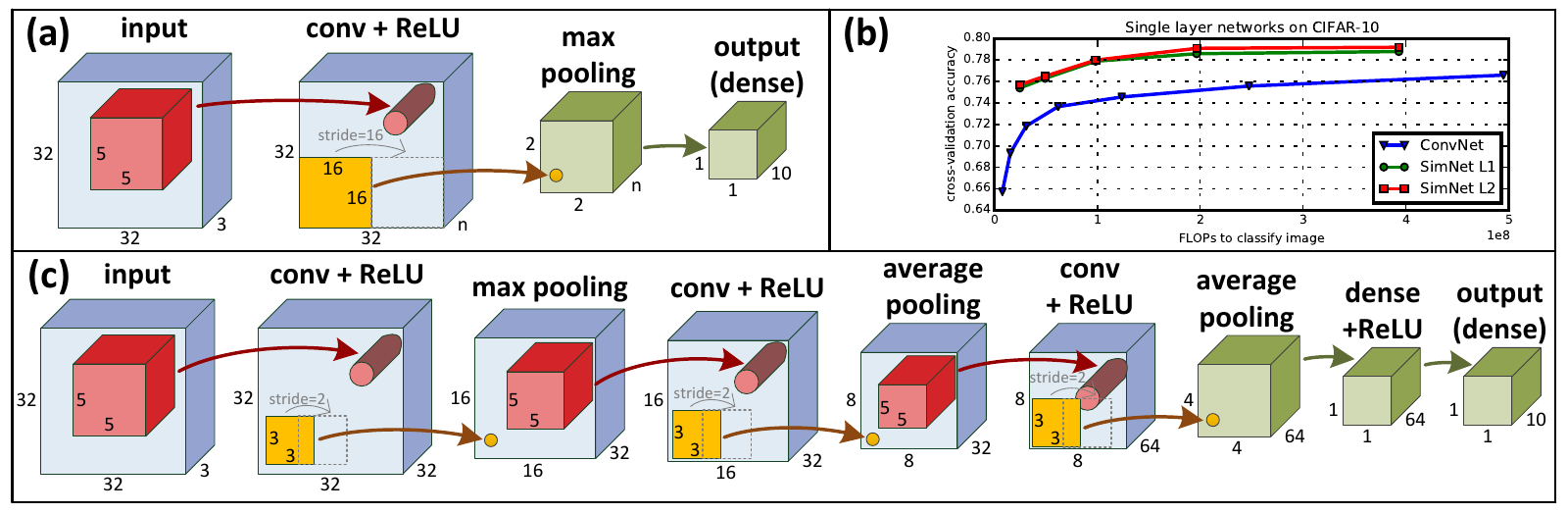}
\caption{\footnotesize{(a)~Single layer ConvNet compared against single layer SimNet on CIFAR-10~~(b)~CIFAR-10 cross-validation accuracies of single-layer networks as a function of the number of floating-point operations required to classify an instance~~(c)~Caffe ConvNet compared against two layer SimNet on CIFAR-10 and SVHN (for CIFAR-100, number of output units increased from 10 to 100).  Best viewed in color.}}
\label{fig:exp_schemes}
\vspace{-3mm}
\end{figure*}

To evaluate the effectiveness of SimNets, we compared them against alternative ConvNets in three experiments of increasing complexity.  In the first experiment, we ran a single layer SimNet against an equivalent single layer ConvNet, and studied the effect of model size (number of convolution/similarity channels) on the accuracy of the two networks.  In a second experiment, we compared compact two layer SimNets against the best performing publicly available ConvNet we are aware of that has comparable complexity.  In the third and final experiment, we constructed a large three layer SimNet designed to compete against state of the art ConvNets.  Our experiments demonstrate that SimNets are significantly more accurate than ConvNets when networks are constrained to be compact, i.e. when computational load at run-time is limited.  This complies with our theoretical analysis in sec.~\ref{sec:simnet_mlp}, which shows that weighted $\ell_p$ similarity exhibits an expressive power that goes beyond kernel machines, whereas linear similarity (the case associated with ConvNets) is fully captured by the Exponential kernel.  Asymptotically as the dimension increases, even a simple kernel machine becomes expressive enough for a given problem, and more elaborate expressiveness may actually be a burden, as it aggravates overfitting.  Nonetheless, we see in our experiments that with proper regularization, large-scale SimNets achieve accuracies comparable to state of the art ConvNets.
\subsection{Experimental details}
The datasets used in our experiments are CIFAR-10 and CIFAR-100~(\cite{krizhevsky2009learning}), as well as SVHN~(\cite{netzer2011reading}).  These three datasets together form an image recognition benchmark that is diverse and challenging on one hand, yet simple enough to enable granular controlled experiments such as those needed to evaluate a new architecture.  All datasets consist of 32x32 color images.  SVHN (Street View House Numbers) represents a rather simple classification benchmark, where various methods are known to produce near-human accuracies.  It contains approximately 600K images for training and 26K images for testing, partitioned into 10 categories that correspond to the digits 0 through 9.  CIFAR-100 contains 50K images for training and 10K images for testing, equally partitioned into 100 categories.  With a relatively large number of categories, and only a few hundred training examples per class, CIFAR-100 represents a challenging classification task.  CIFAR-10 contains 50K images for training and 10K images for testing, equally partitioned into 10 categories.  It brings forth a balanced trade-off between the simplicity of SVHN and the complexity of CIFAR-100, and accordingly served as the central dataset throughout our experiments.  Namely, all cross-validations were carried out on CIFAR-10 (with 10K training images held out for validation), with SVHN and CIFAR-100 used for final evaluation only.  In terms of implementation, we have integrated SimNets into Caffe toolbox~(\cite{jia2014caffe}), with the aim of making our code publicly available in the near future.

In all our experiments, we trained both SimNets and ConvNets by minimizing softmax loss using SGD with Nesterov acceleration~(\cite{sutskever2013importance}).  Batch size, momentum, weight decay and learning rate were chosen through cross-validation, though we observed, at least for the case of SimNets, that the following choices consistently produced good results: batch size 128, momentum 0.9, weight decay 0.0001 and learning rate 0.01 decreasing by a factor of 10 after 200 and 250 epochs (out of 300 total).  Unlike ConvNets which are mostly initialized randomly nowadays~(\cite{krizhevsky2012imagenet}), SimNets are naturally pre-trained using statistical estimation methods (sec.~\ref{sec:pre_train}).  For computational efficiency, we implemented stochastic versions of these algorithms.  Unless otherwise stated, all reported SimNet results were obtained using its pre-training scheme.
\subsection{Single layer SimNet}
As an initial experiment we compared a single layer SimNet, i.e. a SimNet MLPConv with whitened $\ell_p$ similarity (conv$\to\ell_p$-sim), to an equivalent single layer ConvNet defined for this purpose.  We chose to design the ConvNet in accordance with the prescription given by Coates et al. in their study of single layer networks~(\cite{coates2011analysis}).  The resulting network is illustrated in fig.~\ref{fig:exp_schemes}(a).  As can be seen, it includes a single convolutional layer with 5x5 receptive field and ReLU activation, followed by max pooling over quadrants and dense linear classification.  To align the SimNet with this structure, we applied the whitened similarity to patches with spatial size 5x5, and since these have relatively low dimension already (75), we did not reduce it further during whitening.

To compare the networks as they vary in size (and run-time complexity), we set the number of convolution/similarity channels (denoted $n$ in fig.~\ref{fig:exp_schemes}(a) and fig.~\ref{fig:arch_schemes}(c)) to 50, 100, 200, 400 and 800.  Since the ConvNet requires less computations for a given number of channels, we also tried it with 1600 and 3200 channels.  CIFAR-10 cross-validation accuracies produced by the ConvNet, the SimNet with $\ell_1$ similarity, and the SimNet with $\ell_2$ similarity, are plotted in fig.~\ref{fig:exp_schemes}(b) against the number of FLOPs (floating-point operations) required to classify an image \footnote{In this paper, we consider FLOPs to be a measure of computational complexity.  We do not compare actual run-times, as our implementation of SimNets is relatively na\"{\i}ve, not nearly as efficient as the highly optimized ConvNet code that comes built-in to Caffe.  One may argue that like Caffe, many other hardware or software platforms are specifically designed for convolutions, and therefore ConvNets have a computational edge over SimNets.  While this is true for some off-the-shelf systems, our goal in this paper is to address inherent algorithmic complexities, not specific platforms currently in the market.} \footnote{To circumvent the computational price of $\exp$ and $\log$ functions included in SimNets, we used approximations that require up to 10 FLOPs per operation.  The resulting degradation in accuracy is marginal.}.  As can be seen, for a given computational budget, the accuracies of $\ell_1$ and $\ell_2$ SimNets are comparable, whereas the ConvNet falls significantly behind.
\subsection{Two layer SimNet}
The purpose of this second experiment was to compare SimNets against the best publicly available compact ConvNet we could find.  We are interested in a clean SimNet vs. ConvNet architectural comparison, and thus did not include in the experiment model compression techniques such as those listed in sec.~\ref{sec:intro} (e.g. FitNets~\cite{Romero:2014tg}), which may be applied to both architectures.  An additional reason to exclude these techniques, as well as other works dealing with compact ConvNets (e.g.~\cite{finn2014learning,zhengcompact}), is that all results they report relate to networks that are significantly larger than those we are interested in evaluating, in many cases too large to fit a real-time mobile application.  With the stated purpose of this experiment being a comparison against an off-the-shelf ConvNet that was not altered by us, we eventually chose to work against the compact CIFAR-10 ConvNet that comes built-in to Caffe, the structure of which is illustrated in fig.~\ref{fig:exp_schemes}(c).  As the figure shows, the network includes three 5x5 convolutions, each followed by ReLU activation and pooling.  Two dense linear layers (separated by ReLU) map the last convolutional layer into network outputs (class scores).  The SimNet to which we compared Caffe ConvNet is a two layer network that follows the general structure outlined in fig.~\ref{fig:arch_schemes}(d), with $\ell_2$ similarity and architectural choices taken to maximize the alignment with Caffe ConvNet: 5x5 receptive field and 32 channels in the first similarity layer, 5x5 receptive field and 64 channels in the second similarity layer, and MEX pooling between the similarities fixed to 3x3 max pooling with stride 2.  

\begin{table}
\begin{center}
\begin{tabular}{ccccc}
\hline
\multicolumn{1}{|c||}{\textbf{Network}} & \multicolumn{1}{c|}{\textbf{Acc.} (\%)} & \multicolumn{1}{c|}{\textbf{FLOP}} & \multicolumn{1}{c|}{\textbf{Param.}} \\ 
\hline
\hline
\multicolumn{4}{|c|}{\emph{\textbf{CIFAR-10}}} \\ 
\hline
\multicolumn{1}{|c||}{Caffe ConvNet} & \multicolumn{1}{c|}{81.1} & \multicolumn{1}{c|}{24.8M} & \multicolumn{1}{c|}{145.6K} \\ 
\hline
\multicolumn{1}{|c||}{Two layer SimNet} & \multicolumn{1}{c|}{85.5} & \multicolumn{1}{c|}{14.2M} & \multicolumn{1}{c|}{64.6K} \\ 
\hline
\hline
\multicolumn{4}{|c|}{\emph{\textbf{SVHN}}} \\ 
\hline
\multicolumn{1}{|c||}{Caffe ConvNet} & \multicolumn{1}{c|}{94} & \multicolumn{1}{c|}{24.8M} & \multicolumn{1}{c|}{145.6K} \\ 
\hline
\multicolumn{1}{|c||}{Two layer SimNet} & \multicolumn{1}{c|}{93.8} & \multicolumn{1}{c|}{14.2M} & \multicolumn{1}{c|}{64.6K} \\ 
\hline
\hline
\multicolumn{4}{|c|}{\emph{\textbf{CIFAR-100}}} \\ 
\hline
\multicolumn{1}{|c||}{Caffe ConvNet} & \multicolumn{1}{c|}{52.4} & \multicolumn{1}{c|}{24.8M} & \multicolumn{1}{c|}{151.4K} \\ 
\hline
\multicolumn{1}{|c||}{Two layer SimNet} & \multicolumn{1}{c|}{54.6} & \multicolumn{1}{c|}{14.6M} & \multicolumn{1}{c|}{70.3K} \\ 
\hline
\multicolumn{1}{l}{} & \multicolumn{1}{l}{}               
\end{tabular}
\vspace{-2mm}
\caption {Two layer SimNet vs. Caffe ConvNet on CIFAR-10, SVHN and CIFAR-100~--~comparison of test accuracies, number of floating-point operations required to classify an image, and number of learned parameters.}
\label{tab:simnet2_results}
\vspace{-5mm}
\end{center}
\end{table}

The networks were initially evaluated on CIFAR-10.  Training hyper-parameters for the SimNet were configured via cross-validation, whereas for Caffe ConvNet we used the values that come built-in to Caffe.  After measuring CIFAR-10 test accuracies, the same settings (network architectures and training hyper-parameters) were used to evaluate test accuracies on SVHN.  For evaluation of test accuracies on CIFAR-100, we again used the exact same settings as in CIFAR-10, but this time increased the number of output channels in both networks from 10 to 100.  The results of this experiment are summarized in table~\ref{tab:simnet2_results}.  As can be seen, the SimNet is roughly twice as efficient as Caffe ConvNet, yet achieves significantly higher accuracies on the more challenging benchmarks (CIFAR-10 and CIFAR-100).  On SVHN accuracies are comparable, the reason being that in this simple benchmark classification error is dominated by overfit, to which the enhanced expressiveness of SimNets does not contribute.
\subsection{Three layer SimNet}
In the previous experiments we have seen that SimNets are more accurate than ConvNets when networks are constrained to be compact, i.e. when classification run-time is limited.  In such a setting, the lower approximation error of SimNets plays an important role.  In contrast, when networks are over-specified (i.e. are much larger than necessary in order to model the problem at hand)~--~standard practice for achieving state of the art accuracy, the approximation error is virtually zero, and the advantage of the SimNet architecture fades.  Moreover, the additional expressive power of SimNets could actually be a burden, as additional regularization for controlling overfit would be required.  It is therefore of interest to explore the ability of SimNets to reach state of the art accuracy with over-specified networks.  This is the aim of our third and final experiment, carried out on CIFAR-10. 

In this experiment we used a three layer SimNet as described in fig.~\ref{fig:arch_schemes}(d), with the following architectural choices (determined via cross-validation): $\ell_2$ similarities; 192 similarity channels in all three layers with receptive field sizes 5x5, 5x5 and 3x3 (respectively); max pooling after layer 1, average pooling after layer 2, in both cases pooling windows are 3x3 in size with stride 2 between them.  We trained the network with basic data augmentation, and regularized using multiplicative Gaussian noise\footnote{This regularization technique was shown to be more effective than dropout~(\cite{srivastava2014dropout}), and better suits the nature of SimNets (zeroing out an input coordinate does not neutralize its effect on $\ell_p$ similarity).} in conv$\to\ell_p$-sim layers.  We did not make use of ensembles~(\cite{ciresan2012multi}) or aggressive data augmentation that includes rescaling images~(\cite{graham2014fractional}).  These practices are known to improve accuracy, but are orthogonal to the SimNet vs. ConvNet distinction.  We did not include them in our study in order to facilitate a simpler comparison between the two architectures.  Table~\ref{tab:simnet3_results} draws a comparison between the test accuracy reached by the SimNet and reported state of the art results that did not make use of ensembles or aggressive data augmentation.  As the table shows, SimNets compare to state of the art ConvNets, even in the over-specified setting.

As a final sanity check, we compared extremely compact versions of our three layer SimNet and Network in Network (NiN,~\cite{lin2013network}) \footnote{We chose to work against NiN since it bears an architectural resemblance to our SimNet, thus it was clear how both networks can be made compact in an analogous way.}.  Specifically, we changed the number of channels in all layers of both networks to 10, and removed dropout (NiN) and multiplicative Gaussian noise (SimNet), leaving all other hyper-parameters intact.  The resulting networks had only 5K parameters each, and required just 3.5M FLOPs to classify an image.  With such limited resources we expect the SimNet to benefit from its inherent expressiveness, and indeed, it outperformed NiN significantly, providing 76.8\% accuracy compared to 72.3\% reached by NiN.

\section{Conclusion}
We presented a deep layered architecture called SimNets that generalizes convolutional neural networks.  The architecture is driven by two operators: (i) the similarity operator, which is a generalization of the inner-product operator on which ConvNets are based, and (ii) the MEX operator, that can realize non-linear activation and pooling, but has additional capabilities that make SimNets a powerful generalization of ConvNets.  An interesting property of the SimNet architecture is that applying its two operators in succession~--~similarity followed by MEX, results in what can be viewed as an artificial neuron in a high-dimensional feature space (sec.~\ref{sec:simnet_mlp}).  This also holds for the more elaborate image processing SimNet incorporating locality, sharing and pooling (sec.~\ref{subsec:simnet_mlpconv}).

\begin{table}
\begin{center}
\begin{tabular}{ccccc}
\hline
\multicolumn{1}{|c||}{\textbf{Method}}  & \multicolumn{1}{c|}{\textbf{Acc.} (\%)} \\ 
\hline
\hline
\multicolumn{1}{|c||}{Network in Network~(\cite{lin2013network})} & \multicolumn{1}{c|}{91.19} \\ 
\hline
\multicolumn{1}{|c||}{Deeply Supervised Nets~(\cite{lee2014deeply})} & \multicolumn{1}{c|}{92.03} \\ 
\hline
\multicolumn{1}{|c||}{Highway Network~(\cite{Srivastava:2015uq})} & \multicolumn{1}{c|}{92.4} \\ 
\hline
\multicolumn{1}{|c||}{ALL-CNN~(\cite{springenberg2014striving})} & \multicolumn{1}{c|}{92.75} \\ 
\hline
\multicolumn{1}{|c||}{Three layer SimNet} & \multicolumn{1}{c|}{92.18} \\ 
\hline
\multicolumn{1}{l}{} & \multicolumn{1}{l}{}                   
\end{tabular}
\vspace{-2mm}
\caption {Three layer SimNet vs. state of the art ConvNets on CIFAR-10 (ensemble and aggressive data augmentation methods excluded) -- comparison of test accuracies.}
\label{tab:simnet3_results}
\vspace{-5mm}
\end{center}
\end{table}

The feature spaces realized by SimNets depend on the choice of similarity type: linear or $\ell_p$ with/without weights.  We have shown that the simplest setting using linear similarity (corresponding to regular convolution) realizes the feature space of the Exponential kernel, while $\ell_p$ settings realize feature spaces of more powerful kernels (Generalized Gaussian, which includes as special cases RBF and Laplacian), or even dynamically learned feature spaces (Generalized Multiple Kernel Learning).  These observations suggest that SimNets, when equipped with $\ell_p$ similarity, have higher abstraction level than ConvNets, which correspond to linear similarity.

We argue that a higher abstraction level for the basic network building blocks carries with it the advantage of obtaining higher accuracies with small networks, an important trait for mobile and real-time applications.  Through a detailed set of experiments we validated the conjecture of higher accuracy for small networks, and we have also shown that SimNets can achieve state of the art accuracy in large-scale settings where computational efficiency is not a concern (and thus the higher abstraction per given network size is not an advantage).

Finally, the SimNet architecture is endowed with a natural pre-training scheme based on unlabeled data.  Besides its aid in training, the scheme also has the potential of determining the number of channels in hidden layers based on statistical analysis of patterns generated in previous layers.  This implies that the structure of SimNets can potentially be determined automatically based on (unlabeled) training data.  Future work includes a study of this capability, and more generally, further analysis of probabilistic properties of SimNets and unsupervised/supervised algorithms derived thereof.  

\ifdefined\CAMREADY
	\subsubsection*{Acknowledgments}
	We thank Ronen Tamari for his dedicated contribution to the experiments.  The work is partly funded by Intel grant ICRI-CI 9-2012-6133 and ISF grant 1790/12.  Nadav Cohen is supported by a Google Fellowship in Machine Learning.
\fi

\subsubsection*{References}
\small{
\bibliographystyle{plainnat}
\bibliography{refs.bib}
}

\end{document}